\documentclass[11pt]{article}
\usepackage{acl2015}
\usepackage{times}
\usepackage{url}
\usepackage{latexsym}
\usepackage{graphicx}
\usepackage{booktabs}
\usepackage[margin=1in]{geometry}
\usepackage{longtable}
\usepackage{listings} 

\title{Machine Generated Product Advertisements: Benchmarking LLMs Against Human Performance}

\author{Sanjukta Ghosh \\
  University at Buffalo \\
  NY, USA \\
  {\tt sdgsanjuktaghosh@gmail.com} }

\date{}

\begin{document}
\maketitle
\begin{abstract}
This study compares the performance of AI-generated and human-written product descriptions using a multifaceted evaluation model. We analyze descriptions for 100 products generated by four AI models (Gemma 2B, LLAMA, GPT2, and ChatGPT 4) with and without sample descriptions, against human-written descriptions. Our evaluation metrics include sentiment, readability, persuasiveness, Search Engine Optimization(SEO), clarity, emotional appeal, and call-to-action effectiveness. The results indicate that ChatGPT 4 performs the best. In contrast, other models demonstrate significant shortcomings, producing incoherent and illogical output that lacks logical structure and contextual relevance. These models struggle to maintain focus on the product being described, resulting in disjointed sentences that do not convey meaningful information. This research provides insights into the current capabilities and limitations of AI in the creation of content for e-Commerce.
\end{abstract}

\section{Introduction}
\textbf{1.1 Background:}
The rise of artificial intelligence in content creation has revolutionized the landscape of digital marketing, offering new possibilities for efficient and scalable content production. AI tools are increasingly being used to generate various types of content, from product descriptions to blog posts and social media updates. This technological advancement has sparked a debate on the effectiveness and quality of AI-generated content compared to human-written content.
Product descriptions play a crucial role in e-Commerce, serving as a virtual sales pitch that bridges the gap between products and potential customers. Well-crafted product descriptions not only inform customers about product features and benefits, but also help in connecting with the ideal customer, showcasing value, and boosting search engine visibility. In the competitive e-commerce landscape, compelling product descriptions can significantly impact conversion rates, customer loyalty, and overall sales performance.
The current debate on AI vs. human-written content centers around several key aspects, including creativity, emotional engagement, accuracy, and scalability. Although AI excels in efficiency and consistency, human writers are praised for their ability to bring creativity, nuance, and emotional depth to content. This ongoing discussion raises important questions about the future of content creation and the potential for a hybrid approach that leverages the strengths of both AI and human writers.

\textbf{1.2 Research Objectives:}
This study aims to provide a comprehensive analysis of AI-generated and human-written product descriptions, with several specific objectives. The primary goal is to conduct a thorough comparison between AI-generated and human-written product descriptions using a diverse set of performance metrics. These metrics include sentiment analysis, readability, persuasiveness, SEO optimization, clarity, emotional appeal, and call-to-action effectiveness. By evaluating content across these dimensions, we aim to provide a nuanced understanding of the strengths and weaknesses of each approach in creating effective product descriptions.
Furthermore, we seek to assess and compare the capabilities of various AI models, specifically Gemma 2B \cite{gemma2b}, LLAMA \cite{llama}, GPT2 \cite{gpt2}, and ChatGPT 4 \cite{chatgpt4}, in the generation of product descriptions. This evaluation will help identify which AI models excel in particular aspects of content creation and where they may fall short compared to human-written content or other AI models. This analysis will provide valuable insights into the current state of AI technology in natural language generation for e-Commerce applications.
An important aspect of our research is to determine how the provision of sample descriptions affects the quality of AI-generated content. We aim to analyze whether giving AI models examples of well-written product descriptions improves their output across our evaluation metrics. This objective will help us to understand the potential for fine-tuning AI models and the importance of high-quality training data to improve AI performance in content generation tasks.
Through these objectives, this study seeks to contribute to the growing body of knowledge on AI-generated content and its applications in e-commerce, providing insights that can guide future developments in this rapidly evolving field.

\textbf{1.3 Significance of the study:}
This research holds significant implications for e-Commerce content strategies, AI development, and digital marketing practices.
In terms of e-Commerce content strategies, our study provides valuable insights for businesses aiming to optimize their product descriptions. By comparing AI-generated and human-written content across multiple metrics, we offer guidance on when and how to effectively leverage AI in content creation. This can enable e-Commerce businesses to enhance efficiency, maintain consistency, and potentially reduce costs associated with content production.
From an AI development perspective, the analysis of various AI models—Gemma-2B, LLAMA 3.1-8B, GPT-2, and GPT-4—provides a comprehensive understanding of the current capabilities of AI in generating product descriptions. This study contributes to the ongoing discourse on the role of AI in natural language processing and generation, highlighting areas where AI excels and where human input remains crucial \cite{martin2024state,miro2024beyond}.
For digital marketing, the findings offer actionable insights into optimizing the content creation process. Digital marketers can develop hybrid approaches that combine AI efficiency with human creativity. By understanding the strengths and limitations of AI-generated content, marketers can streamline their workflows, resulting in faster production of high-quality, SEO-optimized product descriptions \cite{thompson2024automated}. This optimization can lead to improved search engine rankings, increased organic traffic, and ultimately, higher conversion rates for e-Commerce businesses.

\section{Related Work}
We will now describe some existing work in this domain of research.

\textbf{2.1 AI in Content Generation:}
Text generation AI models have revolutionized content creation by leveraging deep learning techniques to produce human-like text. These models, such as the Generative Pre-trained Transformer(GPT), utilize neural networks trained on vast amounts of text data to understand language patterns and generate coherent, contextually relevant content. Among the key models in this domain, GPT-2 stands out as a transformer-based model renowned for its ability to generate text that is contextually appropriate across a wide range of domains \cite{awan2024text}.
Another notable model is LLAMA (Large Language Model Meta AI), developed by Meta, which is designed to be more efficient and adaptable than earlier models \cite{neha2024exploring}.
Gemma-2B, though less documented in publicly available sources, is presumed to be a variant of Google’s language models, contributing to advancements in natural language generation. Finally, GPT-4, the latest iteration in OpenAI’s GPT series, is recognized for its advanced language understanding and generation capabilities, setting a new benchmark for AI-driven text generation \cite{herbold2023comparison}.
Previous studies comparing AI and human-written content have yielded intriguing results. Recent research has demonstrated that AI-generated content can often match or exceed the quality of human-written text in certain contexts. A large-scale study comparing ChatGPT-generated argumentative essays with human-written ones found that AI-generated essays were rated higher in quality by human experts \cite{herbold2023comparison}. The study revealed several key findings. Firstly, AI models generated significantly higher-quality argumentative essays than human students across all criteria in the scoring rubric. Secondly, ChatGPT-4 outperformed ChatGPT-3 in logical structure, language complexity, vocabulary richness, and text linking. Lastly, writing styles differed significantly between humans and AI models. AI-generated text used more nominalizations and had higher sentence complexity, while human-written text made more use of modal and epistemic constructions.
However, it is important to note that humans still struggle to distinguish between AI-generated and human-written text consistently. Research has shown that people can only differentiate AI-generated text about 53\% of the time, barely above random guessing. This highlights the increasing sophistication of AI text generation models and the challenges in detecting AI-generated content.

\textbf{2.2 Product Description Effectiveness:}
Effective product descriptions are crucial for e-Commerce success, incorporating several key elements to engage customers and drive sales. Customer-focused content that directly addresses the needs and preferences of the ideal customer is essential\cite{flannigan2023product}. A benefit-oriented approach, highlighting how the product solves problems or enhances the customer's life, proves more effective than merely listing features. Storytelling techniques can create an emotional connection with customers, making the product more memorable and appealing \cite{jones2022luxury}. The use of sensory language helps customers imagine using the product, while a clear and scannable format with bullet points, subheadings, and white space enhances readability \cite{duistermaat2024product}. 
Including social proof through customer reviews or testimonials builds trust and credibility. A unique value proposition clearly communicates what sets the product apart from competitors, while accurate and detailed information about features, specifications, and usage helps customers make informed decisions\cite{pimcore2024product}.
Accurate and detailed product information enhances customer satisfaction and fosters loyalty. Transparent and honest descriptions build trust in the brand, influencing purchase decisions and overall brand perception. From an SEO perspective, keyword-rich, unique product descriptions can improve search engine rankings and increase organic traffic \cite{duistermaat2024product}. Comprehensive descriptions mitigate perceived risks, especially for secondhand or high-value products. Detailed product information can differentiate a brand from competitors, potentially increasing market share 
\cite{startwithdata2024incomplete}. Compelling descriptions can increase time spent on product pages, positively impacting user engagement metrics \cite{jones2022luxury}. Lastly, consistent product information across channels helps maintain a cohesive brand image and voice, further strengthening the brand's market position.

\textbf{2.3 Evaluation Metrics in Content Analysis:}
Content analysis in marketing and e-Commerce relies on a diverse set of evaluation metrics to assess the effectiveness of product descriptions and marketing copy. Sentiment analysis plays a crucial role in understanding the emotional tone of marketing content, helping businesses gauge how their message might be perceived by potential customers. Readability is another critical factor, particularly in e-Commerce, where clear and easily understood product information can significantly impact customer engagement and conversion rates. The use of persuasive language in product descriptions is essential for convincing potential buyers of a product's value and benefits. SEO optimization techniques are vital for ensuring that product descriptions are discoverable by search engines, thereby increasing organic traffic to e-Commerce sites. Clarity and emotional appeal in marketing copy work hand in hand to create compelling narratives that resonate with consumers on both rational and emotional levels. Lastly, the importance of an effective call-to-action in product descriptions cannot be overstated, as it guides potential customers towards making a purchase decision. These diverse metrics collectively provide a comprehensive framework for evaluating and improving the quality and effectiveness of marketing content in the digital marketplace.

\section{Methodology}
Our research methodology is highlighted in this section. We will describe the dataset used by us, our process flow, the LLMs used by us for automated advertisement generation and finally, some metrics for a comprehensive evaluation of our approach. 
 \begin{figure*}[h!]
    \centering
    \includegraphics[width=\linewidth]{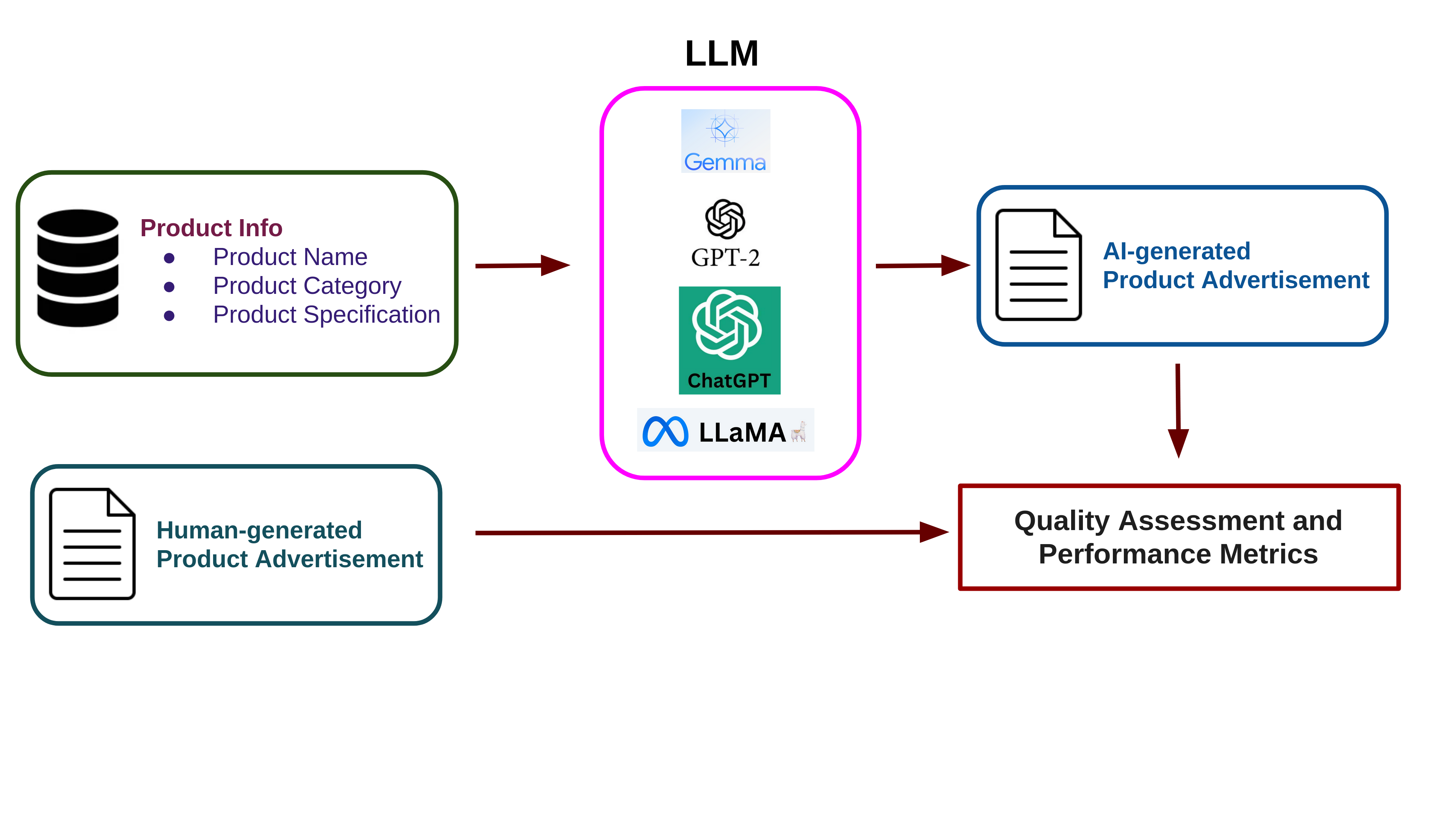}
    \vspace{-20mm}
    \caption{Overview of our proposed approach for the assessment of LLM generated advertisements compared against our human benchmark}
    \label{fig:enter-label}
\end{figure*}

\textbf{3.1 Dataset:}
The dataset for this study comprises 100 products sourced from the recently released Hugging Face dataset "philschmid/amazon-product-descriptions-vlm" \cite{dataset}, which provides a comprehensive collection of product information. Each product entry within the dataset includes essential details such as the product name, product category, and key features or specifications about the product. The human-written product descriptions were directly obtained from this dataset, which offers authentic, real-world examples of product content originally sourced from Amazon.
While the dataset contains product URLs, these were intentionally excluded from the description generation process to maintain research consistency. The research methodology employed a unique approach by utilizing the same 100 products to generate descriptions across different text generation models. This was accomplished both with and without providing sample descriptions, enabling a systematic and controlled comparative analysis across various AI models and human-written content. By maintaining a consistent product set, the study ensures a robust and reliable evaluation of AI-generated product descriptions against their human-written counterparts.

\textbf{3.2 Process Flow Diagram: }
Our approach is outlined in Figure 1. We employ a structured process to generate and evaluate product descriptions using various Large Language Models (LLMs) and compare them with human-generated content. The process flow can be summarized as follows: (1) Data Input: A dataset containing information on 100 products, including product names, categories, and key features, serves as the foundation for both AI-generated and human-written descriptions. (2) AI Description Generation: Four distinct LLMs - Gemma 2B, GPT-2, ChatGPT-4, and LLAMA 3.1 8B Instruct - are used to generate product descriptions based on this input data. (3) Human-Generated Descriptions: The original human-written descriptions from the dataset are retained for comparison. (4) Quality Assessment: Various metrics including sentiment analysis, readability, persuasiveness, SEO optimization, clarity, emotional appeal, and call-to-action effectiveness are calculated for both AI-generated and human-written descriptions to determine the quality of the descriptions.

\textbf{3.3 LLMs and Generation Process:}

This study utilized four LLMs for generating product descriptions, each with unique characteristics and capabilities. Gemma 2B, a lightweight, open-source model from Google with 2 billion parameters, was employed alongside LLAMA, a language model developed by Meta AI. The study also incorporated GPT-2, a transformer-based model known for contextual text generation, and ChatGPT-4, the latest iteration of OpenAI's GPT series. The generation process involved two distinct conditions to evaluate the performance of the models comprehensively. In the first condition, descriptions were generated without samples, using only the product information as input. The second condition provided example description along with the product information to assess the impact of additional context on the generated content. To ensure a fair comparison across all models, consistent parameters and settings were maintained throughout the generation process. These settings were carefully adjusted to optimize performance while preserving consistency, allowing for a robust evaluation of each model's capabilities in producing effective product descriptions.

\textbf{3.4 Evaluation Metrics:}
The evaluation metrics used in this study provide a comprehensive framework for analyzing product descriptions across various dimensions of effectiveness. Sentiment Analysis, utilizing the "distilbert-base-uncased-finetuned-sst-2-english" model \cite{Sanh2019DistilBERTAD} from Huggingface, offers insights into the overall emotional tone of the content, with scores ranging from 0 (very negative) to 1 (very positive). This metric helps to gauge how potential customers might perceive the product emotionally. The Readability metric, based on the Flesch-Kincaid Grade Level formula, estimates the United States school grade level required to understand the text. It considers factor such as sentence length and word complexity, with lower scores indicating easier readability. This metric is crucial for ensuring that product descriptions are accessible to the target audience, as research shows that the average American adult reads at a seventh to ninth grade level.
Persuasiveness is measured by calculating the ratio of predefined persuasive words to the total word count, aiming to quantify the text's potential influence on reader decisions. The SEO Optimization metric assesses how well the description incorporates category-related keywords, which is vital for improving search engine visibility and attracting potential customers.
Clarity is evaluated using the inverse of the average word length, based on the assumption that shorter words contribute to clearer text. This metric helps ensure that product descriptions are concise and easily understood. The Emotional Appeal metric counts the occurrence of predefined emotion-related words, indicating how well the description engages customers on an emotional level.
Lastly, the Call-to-Action (CTA) Effectiveness is measured by counting predefined CTA phrases, assessing how directly the description encourages user action. This metric is particularly important in e-Commerce, where clear calls-to-action can significantly impact conversion rates.
Together, these metrics provide a multi-faceted approach to evaluating product descriptions, covering aspects from basic readability to marketing effectiveness and emotional engagement. This comprehensive analysis allows for a nuanced understanding of how different AI models and human writers perform in creating compelling and effective product content.

\section{Results and Evaluation}
Based on the methodology outlined in the previous section, we conduct an extensive set of experiments. The results are presented in this section along with commentary on the conclusions of our study. 
\begin{table*}[h!]
\centering
\small 
\setlength{\tabcolsep}{3pt} 
\caption{Performance Comparison Across Metrics. Best performing models are in bold for each metric.}
\label{tab:comparison}
\begin{tabular}{@{}lccccccc@{}}
\toprule
Model               & Sentiment & Readability & Persuasiveness & SEO Optimization & Clarity & Emotional Appeal & Call-to-Action \\ \midrule
GPT2                & 0.972    & 6.907    & 0.014         & 0.011           & \textbf{0.218}  & 0.007           & 0.033         \\
ChatGPT4 (manual)   & \textbf{0.996}    & 9.876      & \textbf{0.062}         & \textbf{0.039}           & 0.195  & \textbf{0.013}         & \textbf{0.301}        \\
Gemma               & 0.953    & 9.637      & 0.006         & 0.008           & 0.207  & 0.011           & 0.085         \\
Gemma (Sample)      & 0.954    & 8.712      & 0.007         & 0.008           & 0.208  & 0.011           & 0.160         \\ 
GPT2 (Sample)       & 0.983    &\textbf{7.556}      & 0.023         & 0.007           & 0.200  & 0.009           & 0.019         \\
LLAMA (Sample)      & 0.984    & 8.969      & 0.009         & 0.015           & 0.198  & 0.008           & 0.217        \\
LLAMA               & 0.990    & 9.214      & 0.010         & 0.014           & 0.201  & 0.010           & 0.123         \\ \midrule
Human Generated     & 0.998    & 7.286      & 0.077         & 0.038           & 0.186  & 0.016           & 0.321         \\
\bottomrule
\end{tabular}
\end{table*}

\textbf{4.1 Metric-Specific Findings:}
The results are presented in Table 1, which compares the performance of AI-generated and human-written product descriptions across various metrics. 
The analysis of AI-generated and human-written product descriptions revealed several intriguing patterns and trends across various metrics. In sentiment analysis, all models demonstrated consistently high performance, with scores above 0.95, suggesting that conveying appropriate sentiment in product descriptions is a relatively solved problem for both AI and human writers. Readability scores showed an interesting trend, with most AI models scoring higher than the ideal range of 7-9, while human-generated content fell within this range. This pattern indicates that human writers are better at producing text accessible to a wider audience (7th to 9th-grade level), while AI models tend to generate more complex content that might be less accessible to the target audience. In terms of persuasiveness, a clear trend emerged with human-generated content and ChatGPT4 (manual) significantly outperforming other AI models, falling within the ideal range of 0.06-0.10. This suggests that advanced language models and human writers are more adept at incorporating persuasive elements effectively. SEO optimization showed a similar pattern, with human-generated content and ChatGPT4 (manual) demonstrating a distinct advantage over other AI models, implying that advanced AI models may be approaching human-level performance in applying SEO principles. Clarity scores revealed an interesting pattern, with GPT2 and Gemma achieving slightly higher scores compared to other models and human-generated content. This trend suggests that these AI models are using shorter words in their descriptions compared to more advanced models like ChatGPT4 and human-generated content. Emotional appeal scores showed a gradual improvement from simpler to more advanced AI models, with human-generated content scoring highest, suggesting room for improvement in AI's ability to incorporate emotionally resonant language. Finally, call-to-action effectiveness displayed a clear trend with human-generated content and ChatGPT4 (manual) significantly outperforming other AI models, indicating that advanced AI models are approaching human-level capability in crafting compelling action-oriented content.

\textbf{4.2 Conclusion:}
The analysis of various metrics across AI-generated and human-written product descriptions reveals a nuanced landscape in e-Commerce content creation. Sentiment analysis shows high proficiency across all models, indicating that conveying appropriate sentiment is a solved problem for both AI and human writers. In terms of readability, human-generated content and some AI models like GPT-2 achieve the ideal range, making their descriptions more accessible to a wider audience. Advanced models like ChatGPT-4 and human writers excel in persuasiveness and SEO optimization, demonstrating the potential of AI to match human-level performance in these crucial areas. While simpler AI models showed higher clarity scores due to their use of shorter words, this metric should be interpreted cautiously, as longer words don't necessarily indicate complexity or reduced understandability. Emotional appeal and call-to-action effectiveness still favor human-generated content and advanced AI models like ChatGPT-4, highlighting areas where sophisticated language understanding is crucial. These findings suggest that while AI, particularly advanced models like ChatGPT-4, is rapidly approaching human-level performance in many aspects of product description generation, human expertise still holds value, especially in crafting accessible, persuasive, and emotionally resonant content. The results point towards the potential of a hybrid approach, combining the consistency and efficiency of AI with human creativity and strategic insight. This collaboration could revolutionize e-Commerce content creation, producing more effective and engaging product descriptions that leverage the strengths of both AI and human expertise.

\section{Implications of the Study}
For e-Commerce businesses and content creators, AI can efficiently generate product descriptions at scale, offering significant time and resource savings. However, human oversight remains crucial to ensure quality control and maintain brand consistency, as AI-generated content may require fine-tuning to match specific brand voices. Similarly, businesses can leverage AI to streamline content creation processes while still prioritizing the alignment of descriptions with their unique branding strategies.
When it comes to content creation strategies, hybrid approaches that combine AI generation with human editing are expected to become more common. This shift allows content creators to focus more on higher-level strategies and creative direction, while routine tasks are delegated to AI. Additionally, there is an increased emphasis on prompt engineering and AI model fine-tuning, which are becoming essential skills. AI also enables the generation of more personalized product descriptions at scale, catering to diverse audiences effectively. As a result, content teams may need to develop new skills, such as crafting effective prompts and evaluating AI outputs, while resources are reallocated from traditional writing tasks to AI management and quality assurance.

\section{Limitations and Future Work}
\textbf{6.1 Limitations of the Study:}
Sample size considerations, biases in evaluation metrics, limitations in determining overall quality, and lack of context-specific evaluation are important factors to consider when interpreting the results of this study. The analysis of 100 product descriptions, while informative, may not provide a fully representative sample across all product categories, potentially limiting the generalizability of the findings. The chosen evaluation metrics, including sentiment, readability, persuasiveness, and SEO optimization, offer valuable insights but may not capture all aspects contributing to an effective product description. Automated metrics might overlook nuanced elements such as creativity, brand voice consistency, contextual relevance, cultural sensitivity, storytelling elements, subtle humor or wit, and trendiness in language use. Furthermore, while these metrics provide useful data points, they may not be sufficient to definitively assess whether a description is "good enough" for various marketing purposes, given the subjective nature of what constitutes effective content across different target audiences, product types, and marketing strategies. Additionally, the evaluation metrics might not account for industry-specific jargon or writing styles that could be appropriate or effective in certain contexts, further highlighting the complexity of comprehensively assessing product description quality.

\textbf{6.2 Future Research Directions:}
Future research in AI-generated product descriptions should focus on several key areas. Evaluating ChatGPT-4 generated content is crucial, as it has demonstrated strong performance across various metrics. Comparing these descriptions to both human-written content and outputs from other AI models can provide valuable insights into how consumers perceive and react to ChatGPT-4-generated content in practical contexts. Additionally, exploring non-mechanical metrics is essential to move beyond traditional evaluation methods. This could involve conducting user studies to gather data on emotional responses, trust levels, and purchase intent based on different content sources. Developing a scoring system that measures alignment with brand voice and marketing objectives can also assess how well AI-generated content aligns with company goals and principles. Another significant area for exploration is the impact of disclosing AI involvement in content creation, investigating how consumers respond when informed that a product description was generated by an AI model. This research can illuminate the ethical implications and influence on consumer trust associated with AI-generated content. Lastly, studying the long-term performance of AI-generated content by tracking metrics such as conversion rates, customer satisfaction, and SEO rankings over time can evaluate the sustained effectiveness of AI-generated descriptions compared to those written by humans. 

\bibliographystyle{acl}
\bibliography{references}
\onecolumn
\appendix
\section{Appendix}
\subsection{Example prompts and results}
The table below shows some of the results from using various LLMs. The prompt structure used is also listed as follows.

\begin{lstlisting}
prompt = (
    f"Write a product description for the following product:\n"
    f"Product Name: {product_name}\n"
    f"Product Category: {product_category}\n"
    f"About the Product: {about_product}\n\n"
    "Example1:\n"
    "Adorable Iwako Japanese Vehicle & Plane Eraser Set!"
    "Fun and functional erasers perfect for kids, "
    "students, and collectors. These high-quality erasers 
    feature detailed designs of various vehicles and planes. "
    "Great for school, office, or creative projects. Shop now!\n"
    "Avoid using headers like 'Introduce the Product' or
    'Highlight Key Features.' Focus only on the product's benefits "
    "and features in a consumer-friendly tone. Only
    generate description, no unnecessary details.Keep it concise."
    "Product Description:-"
    )

Ideal persuasive score 0.06 and 0.10 
\end{lstlisting}

\begin{longtable}{|p{2.5cm}|p{2.5cm}|p{2.5cm}|p{7cm}|}
\hline
\textbf{Product Name} & \textbf{Category} & \textbf{Description Source} & \textbf{Generated Description} \\ \hline
\endfirsthead
\hline
\textbf{Product Name} & \textbf{Category} & \textbf{Description Source} & \textbf{Description} \\ \hline
\endhead
\hline
\endfoot

Board \& Dice Inuit: The Snow Folk & Toys \& Games | Games \& Accessories | Board Games & AI Generated - GPT2 & Board \& dice inuit (from the tabletop RPG) are an amazing new category of board games that combine the power of dice with the magic of magic cards.
The board game in this category is an open-ended game where players use their dice to solve complex puzzles. Players must solve a simple problem with dice and the dice will be placed on a table in front of the players to determine their answer. Each player has their own dice for the game and can choose from a variety of different dice types, ranging from simple to very complicated. With each dice roll, each player uses their board and dice pool to build their deck of cards, building a deck that includes dice, cards and objects. (truncated for brevity)
 \\ \hline

Board \& Dice Inuit: The Snow Folk & Toys \& Games | Games \& Accessories | Board Games & AI Generated - GPT2 - with Sample & For 2-4 players. Ages 13+. 45 minute playing time.Two full expansions included in the box.Inuit: the snow folk is a card-based strategy Game of drafting and Tableau building. Simple and intuitive open Daft mechanism. Amazing and immersive illustrations
 \\ \hline

Board \& Dice Inuit: The Snow Folk & Toys \& Games | Games \& Accessories | Board Games & AI Generated - LLAMA & In Inuit: 
- Players take turns drafting card to build their snowmen, each with its unique abilities and point value.
- The player with the most points at the end of the game wins.
Key Features:
- Easy to Learn, Hard to Master
  This game requires a strategic approach to building your village and adapting to the changing environment.
  Players must balance short-term needs with long-term goals, all while navigating the challenges of building a snow community.
- Immersive and Interactive
    The unique Snowstorm mechanism adds an element of unpredictability, forcing players to adapt and make tough decisions.
    This mechanic creates a thrilling experience, as players navigate the ever-changing Arctic landscape.
This game combines simple gameplay with deep strategic thinking, appealing to a wide range of players, from families to casual board game enthusiasts. Each game takes about 20-30 minutes to play, so it's perfect \\ \hline

Board \& Dice Inuit: The Snow Folk & Toys \& Games | Games \& Accessories | Board Games & AI Generated - LLAMA - with Sample &  "The Snow Folks" strategy game. Combine drafting skills with tableau-building strategies in this immersive card experience. For two or four player, perfect strategy and card enthusiasts. Included in package: Two complete expansions to add depth and variety. Explore Inuptiat Arctic culture through detailed, beautiful illustrations. Play time is approximately thirty to sixty minutes, making it a perfect fit for any gaming session.  Great fun for strategy fans, collectors, families and friends. "  \\ \hline

Board \& Dice Inuit: The Snow Folk & Toys \& Games | Games \& Accessories | Board Games & AI Generated - Gemma & Inuit: Snowfolk is an abstract, abstract strategy game for 3-5 players, with a focus on thematically and strategically balancing the game’s 500+ cards. The game is played over a series of 6 rounds, and each round is broken down into 8 phases. Each player has a hand of up to 7 cards, which are then shuffled and dealt out. Players then take it in turns to play a single card from their hand, then resolve the effects of that card. Once a player’snowfolk’ is resolved, they are removed from the tableau, leaving the player with 0 or 9 snowfolk. If a playersnowfolkis removed, that player loses the round. Snowflakes are also removed if they’rein play. A game of Inuit is won when a certain number of rounds have been played, or if a snowflake is removed.  \\ \hline

Board \& Dice Inuit: The Snow Folk & Toys \& Games | Games \& Accessories | Board Games & AI Generated - ChatGPT4 manual & Inuit: The Snow Folk is a captivating card-based strategy game for 2-4 players, ages 13 and up, with a 45-minute playing time. Players engage in drafting and tableau-building, using simple and intuitive mechanics to create their own strategies. The game includes two full expansions, enhancing replayability, and features stunning, immersive illustrations that bring the theme to life. Perfect for fans of strategic gameplay looking for an engaging and visually appealing experience.\\ \hline

Board \& Dice Inuit: The Snow Folk & Toys \& Games | Games \& Accessories | Board Games & Human Generated & Experience the captivating world of Inuit: The Snow Folk! This award-winning board game immerses players in a thrilling strategy game of resource management and cultural preservation.  Perfect for families and experienced gamers alike.  Ages 10+. \\ \hline

\end{longtable}

\end{document}